# Fully automated landmarking and facial segmentation on 3D photographs

Bo Berends[a], Freek Bielevelt[a], Ruud Schreurs[a,b], Shankeeth Vinayahalingam[c], Thomas Maal[a], Guido de Jong[a]

[a]*Radboudumc 3D Lab, Radboud University Medical Center, Nijmegen, The Netherlands*

[b]*Department of Oral and Maxillofacial Surgery, Amsterdam University Medical Center (UMC), Amsterdam, The Netherlands.*

[c]*Department of Oral and Maxillofacial Surgery, Radboud University Medical Center Nijmegen, Nijmegen, The Netherlands.*

**Abstract**
Three-dimensional facial stereophotogrammetry provides a detailed representation of craniofacial soft tissue without the use of ionizing radiation. While manual annotation of landmarks serves as the current gold standard for cephalometric analysis, it is a time-consuming process and is prone to human error. The aim in this study was to develop and evaluate an automated cephalometric annotation method using a deep learning-based approach. Ten landmarks were manually annotated on 2897 3D facial photographs by a single observer. The annotation process was repeated by the first observer, a second observer, and a third observer on 50 randomly selected 3D photos to assess intra-observer and inter-observer variability. The automated landmarking workflow involved two successive DiffusionNet models and additional algorithms for facial segmentation. The dataset was randomly divided into a training (85%) and test (15%) dataset. The training dataset was used to train the deep learning networks, whereas the test dataset was used to evaluate the performance of the automated workflow. The landmarks were also annotated using a semi-automatic method on all 3D photographs. The precision of the workflow was evaluated by calculating the Euclidean distances between the automated and manual landmarks and compared to the intra-observer and inter-observer variability of manual annotation and the semi-automated landmarking method. The workflow was successful in 98.6% of all test cases. The deep learning-based landmarking method achieved precise and consistent landmark annotation. The mean precision of 1.69 ± 1.15 mm was comparable to the inter-observer variability (1.31 ± 0.91 mm) of manual annotation. The Euclidean distance between the automated and manual landmarks was within 2 mm in 69%. Automated landmark annotation on 3D photographs was achieved with the DiffusionNet-based approach. The proposed method allows quantitative analysis of large datasets and may be used in diagnosis, follow-up, and virtual surgical planning.

**Introduction**
The fields of genetics, orthodontics, craniomaxillofacial surgery, and plastic surgery have greatly benefitted from advances in imaging technology, particularly in three-dimensional (3D) imaging. Three-dimensional stereophotogrammetry has gained popularity in these fields since it can capture a detailed and accurate representation of craniofacial soft tissue without the use of ionizing radiation. (Dindaroğlu et al., 2016; Heike et al., 2010; Liu et al., 2021)

Cephalometric analysis can be performed on 3D stereophotographs to extract information about the position of individual landmarks or distances and angles between several landmarks, with the purpose of objectifying clinical observations (Serafin et al., 2023). Despite being a commonly used diagnostic tool in the craniofacial region, landmarking often remains a manual task that is time-consuming, prone to observer variability, and affected by observer fatigue and skill level. (Park et al., 2019; Stewart et al., 2008) Therefore, there has been a growing interest in using artificial intelligence (AI), such as deep learning and machine learning algorithms to automate the landmark identification process.

*Table 1: Dataset distribution with the percentage of the total dataset.*

| Dataset | Train | | Test | | Total | |
|---|---|---|---|---|---|---|
| | n | % | n | % | n | % |
| Headspace | 1053 | 36.3% | 192 | 6.6% | 1245 | 43.0% |
| Radboudumc control | 974 | 33.6% | 165 | 5.7% | 1139 | 39.3% |
| Radboudumc patient | 436 | 15.1% | 77 | 2.7% | 513 | 17.7% |
| **Total** | 2463 | 85% | 434 | 15% | 2897 | 100% |

Several studies have described the use of deep learning algorithms for the automation of hard-tissue landmark extraction for cephalometric analysis (Guo et al., 2020; Serafin et al., 2023). Studies that include soft-tissue landmarks utilize (projective) 2D imaging, are pose dependent, or require manual input (Manal et al., 2019; White et al., 2019). Since only a limited number of studies were performed on the automated extraction of facial soft-tissue landmarks from 3D photographs, this study aimed to develop and validate an automated approach for the extraction of soft-tissue facial landmarks from 3D photographs using deep learning (Baksi et al., 2021; Guo et al., 2013).

**Material and methods**

Data acquisition

In total, 3188 3D facial photographs were collected from two databases: the Headspace database (n=1519) (Dai et al., 2020; Pears et al., 2018) and the Radboudumc's longitudinal database (n=1669). The Radboudumc's data consisted of healthy volunteers (n=1153) and Oral and Maxillofacial Surgery patients (n=516). The Radboudumc dataset was collected in accordance with the World Medical Association Declaration of Helsinki on medical research ethics. The following ethical approvals and waivers were used: CMO 2007/163; ARB NL 17934.091.07; RUMC CMO 2019-5793. All data were captured using 3dMD's 5-pod 3dMDhead systems (3dMDCranial, 3dMD, Atlanta, Georgia USA). Exclusion criteria were large gaps within the mesh, stitching errors, excessive facial hair interfering with the facial landmarks, meshes that lacked texture (color information), and mesh-texture mismatches. An overview of the data is presented in Table 1.

Data annotation

The 3D photographs were manually annotated by a single observer using the 3DMedX® software (v1.2.29.0, 3D Lab Radboudumc, Nijmegen, The Netherlands; details can be found at https://3dmedx.nl). The following ten cephalometric facial landmarks were annotated: exocanthions, endocanthions, nasion, nose tip, alares, and cheilions. The texture of the 3D photographs was used in the annotation process as a visual cue. Manual annotation was repeated on 50 randomly selected 3D photos by the first observer, a second observer, and a third observer to assess the intra-observer and inter-observer variability.

Automated landmarking workflow

The automated landmarking workflow was developed for the same cephalometric landmarks (exocanthions, endocanthions, nasion, nose tip, alares, and cheilions). It consisted of four main steps: 1) rough prediction of landmarks using an initial

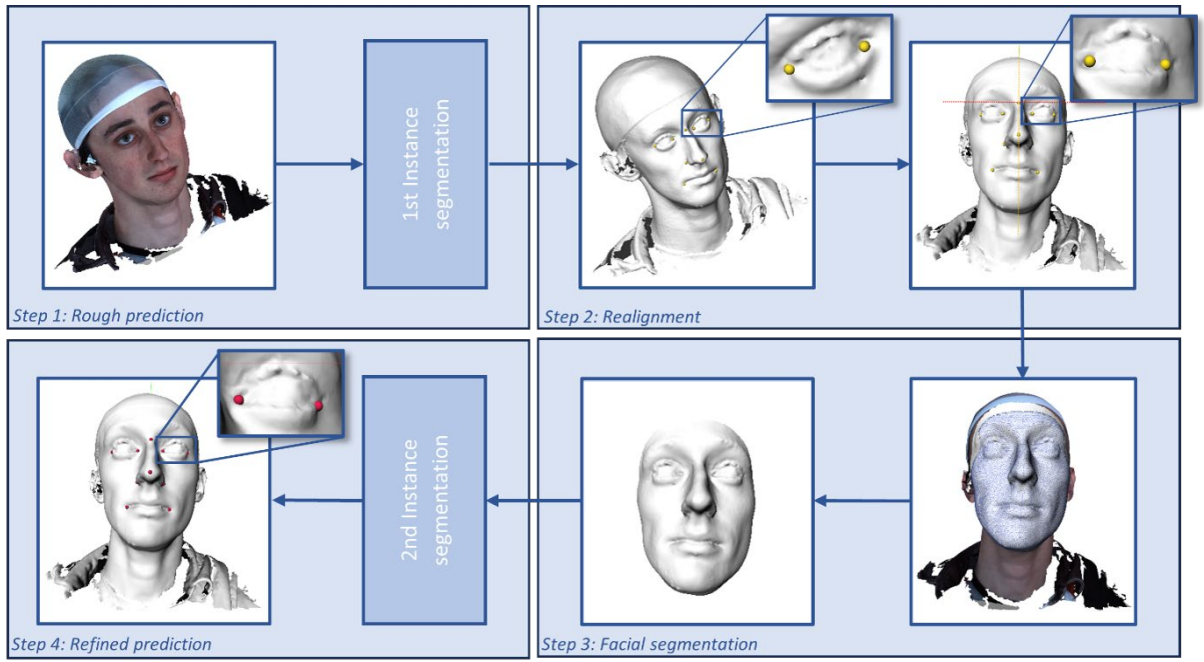

*Figure 1: Automated landmarking workflow. Step 1: First instance segmentation task for rough landmark prediction. Step 2: Realignment of the meshes using the roughly predicted landmarks. Step 3: Facial region segmentation (white) using MeshMonk (blue wireframe). Step 4: Second instance segmentation task for refined landmark prediction.*

DiffusionNet (Sharp et al., 2022) on the original meshes; 2) realignment of the meshes based on the roughly predicted landmarks; 3) segmentation of the facial region through fitting of a template facial mesh using a morphable model; 4) refined landmark prediction on the segmented meshes using a final DiffusionNet. The DiffusionNet models used spatial features only and did not use texture information for the automated landmarking task. An overview of the workflow can be seen in Figure 1.

Training

The data were randomly divided into two sets, 85% for training and 15% for testing of the DiffusionNet models. As a data augmentation step, the 3D meshes from the training dataset were mirrored over the YZ plane to double the number of scans available for training. No validation set was used during training.

**Step 1: Rough prediction of landmarks**

A DiffusionNet, a state-of-the-art and robust deep learning network for 3D surfaces, was utilized for initial prediction of the exocanthions, endocanthions, nasion, nose tip, alares, and cheilions as visualized in Figure 2. (Sharp et al., 2022)

*Preprocessing*

To speed up the training process, each mesh was downsampled to a maximum of 25.000 vertices (Garland & Heckbert, 1997). Subsequently, a mask was applied, assigning a value of 1 to all vertices located within 5 mm Euclidean distance to the manually annotated landmarks and a value of 0 to the remaining vertices.

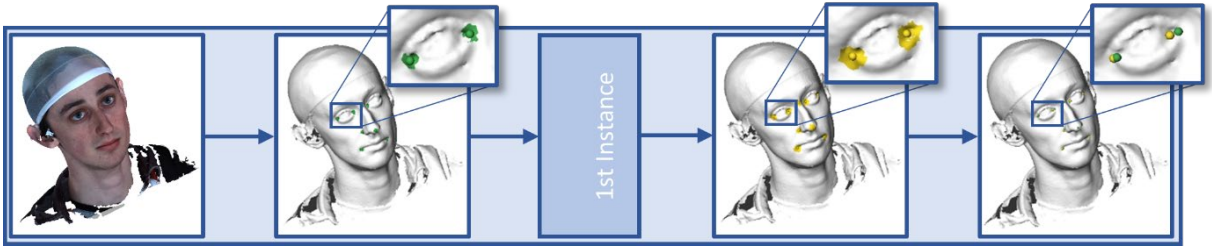

*Figure 2: First instance segmentation task. The manually annotated landmarks (spheres) and corresponding masks are visualized in green. The green areas represent the vertices within 5 mm of the manually annotated landmark. The roughly predicted landmarks are visualized in yellow. The yellow area represents the positively predicted vertices out of which the rough landmarks will be calculated.*

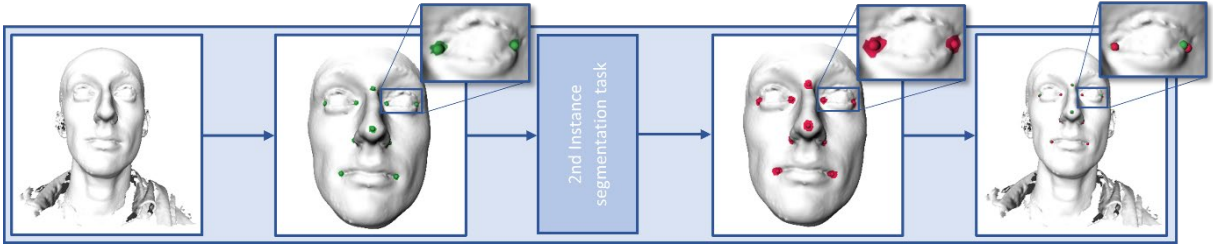

*Figure 3: Second instance segmentation task. The manually annotated landmarks and corresponding masks that were used for training are visualized in green. The green areas represent the vertices within 3.5 mm of the manually annotated landmark. The positively predicted vertices and the calculated refined predicted landmarks are visualized in red.*

### Configuration for the first instance segmentation task (DiffusionNet)

Six output channels were configured for the first instance segmentation task. The two midsagittal landmarks (nasion and nose tip) were assigned an individual channel. The four bilateral landmark pairs were assigned to the four remaining channels. The DiffusionNet model was configured with a C-width (internal dimension) of 256, an MLP (multilayer perceptron layer size) of 256 by 256, and an N-block (number of repeating DiffusionNet blocks) of 12. The network used an Adam optimizer with a Cosine Annealing learning rate of 2 x $10^{-5}$ and a $T_{max}$ of 50 epochs. Furthermore, a binary cross-entropy loss and a dropout rate of 0.10 were applied. Since the orientation and position of the included 3D meshes was not fixed, the network was trained with Heat Kernel Signature (HKS) Features of the 3D meshes. The final output layer was linear. The model was implemented in PyTorch on a 24 GB NVIDIA RTX A5000 GPU and trained for 200 epochs.

### Post-processing

After the instance segmentation, the model was used to predict which vertices belonged to each of the configured channels. For the symmetrical landmarks, a 3D clustering algorithm was utilized to distinguish the predicted vertex clusters from each other. Subsequently, a weighted combination of the output values (activations), as well as the locations of each of the vertices that received a non-zero activation value, were used to determine the landmark positions using Equation 1.

$$Location = \frac{\sum(10^{Activations_{1,2,\dots,i}} * coordinate_{1,2,\dots,i})}{\sum 10^{Activations_{1,2,\dots,i}}} \quad (1)$$

A plane was formed by connecting the predicted nasion, nose tip, and cheilion midpoint to establish if the bilateral landmarks were on the left or right side of the face using the plane equation.

**Step 2: Realignment**

Based on the rough prediction of the exocanthions, nasion, and cheilions, the 3D meshes were positioned in a reference frame. The nasion was defined as the origin, with the x-axis running parallel to the line connecting both exocanthions and the z-axis parallel to the nasion-cheilion midpoint line (Figure 1).

**Step 3: Facial region segmentation**

The MeshMonk algorithm (White et al., 2019), which utilizes a combination of rigid and non-rigid template matching, was used to segment the facial region on the realigned meshes; the default face template of the algorithm was used. The exocanthions, nose tip, and cheilions were used for the initial registration of the facial template mesh to the 3D meshes. The configuration of the MeshMonk fitting algorithm is given in the Appendix. After fitting, the vertices of the aligned 3D meshes that were located further than 15 mm from the fitted template were removed (Figure 1).

The MeshMonk algorithm can also be utilized for landmark annotation. The ten landmarks were collected using this semi-automatic approach to serve as a reference for the precision of the automated annotation approach. In contrast to the automated approach, the manually annotated landmarks were used for template fitting in the semi-automated approach to comply with the MeshMonk workflow. (White et al., 2019)

**Step 4: Refined landmark prediction**

A second instance segmentation task, using DiffusionNet, was used to predict the landmarks on the realigned and segmented 3D meshes (Figure 3).

*Preprocessing*

In contrast to step 1, the meshes were not downsampled. A mask was created in which the vertices within 3.5 mm of the manually annotated landmarks were assigned value 1 and the other vertices were assigned value 0. Default mesh normalization and scaling were applied as provided by the DiffusionNet package (Sharp et al., 2022).

*Configuration for the second instance segmentation task (DiffusionNet)*

For the second instance segmentation task, ten output channels were configured: each individual landmark was assigned to an individual channel. The DiffusionNet was configured with a C-width of 384, an MLP of 768, and an N-blocks of 12. Compared to the first network, the same optimizer, loss, and drop-out were used. However, this second network was trained with XYZ settings instead of HKS settings as the rotation invariance was no longer present after step 3 (Figure 1). The model was implemented in PyTorch on a 24 GB NVIDIA RTX A5000 GPU and trained for 200 epochs. The final output layer was linear.

*Post-processing*

A weighted combination of the activations, supplemented by the locations of each of the vertices, was again used to determine the final landmark positions.

**Statistical analysis**

Statistical analyses were performed on available patient characteristics to assess differences between the source databases and between the training and test data. To assess the intra-observer and inter-observer variability of the manual annotation method, the Euclidean distances between the landmarks annotated by the different observers were calculated. Descriptive statistics were used to summarize the results. The Euclidean distances between the predicted and the manually annotated landmarks were calculated for every test set to evaluate the performance of automated landmarking; descriptive statistics were used for summarizing the results. This was done for both the rough (initial DiffusionNet) and the refined (final DiffusionNet) predictions. The performance of the automated landmarking workflow was compared to the intra-observer and inter-observer variability of the manual annotation method.

*Table 2: Population characteristics per dataset.*

|  | Age (years) | | | | Gender | | |
| --- | --- | --- | --- | --- | --- | --- | --- |
| **Dataset** | Mean | Std | Min | Max | Male | Female | Transgender |
| Headspace | 35.9 | 17.6 | 2 | 90 | 631 (50.7%) | 613 (49.2%) | 1 (0.1%) |
| Controls | 42.1 | 19.4 | 0 | 90 | 492 (43.2%) | 647 (56.8%) | . |
| Patients | 27.8 | 10.9 | 13 | 69 | 190 (37.0%) | 323 (63.0%) | . |

The Euclidean distances between the manually annotated landmarks and the predictions by the semi-automated MeshMonk method were calculated and compared to the precision of the refined predictions using a one-way repeated measures ANOVA test. A p-value <0.05 was used as a cut-off value for statistical significance.

**Results**

Based on the stated exclusion criteria, 291 3D photographs were excluded, yielding a total of 2897 3D photographs that were used for training and testing of the developed workflow (Table 1). Most of the exclusions were due to the lack of texture information (n=271). The age and gender characteristics are given in Table 2. A statistically significant difference was found for age and gender between the source databases (p<0.001 and p<0.001, respectively). However, there were no statistically significant differences between ages and genders of the training and test splits (p=0.323 and p=0.479, respectively). There were no unknown genders or ages in the test dataset. The training dataset held one transgender case and had five unknown ages.

The intra-observer and interobserver differences of the manual annotation method are summarized in Table 3 and Table 4, respectively. The overall mean intra-observer variability for manual annotation of the ten landmarks was 0.94 ± 0.71 mm; the overall mean interobserver variability was 1.31 ± 0.91 mm.

The initial DiffusionNet showed an average precision of 2.66 ± 2.37 mm, and the complete workflow achieved a precision of 1.69 ± 1.15 mm. The performance of both models is summarized in Table 5. The workflow could be completed for 98.6% of the test data; for six 3D photos (1.4%), one of the rough landmarks required for the consecutive steps could not be predicted by the first DiffusionNet. Upon visual inspection, the six excluded 3D photos contained large gaps and/or substantial amounts of information outside the region of interest, such as clothing or hair. Since the workflow could not be completed, these data sets were excluded from the results.

The precision was within 2 mm for 69% of the refined predicted landmarks, within 3 mm for 89% of the landmarks, and within 4 mm for 96% of the landmarks. Table 6 details the precision within these boundaries for the individual landmarks. The exocanthions and alares were found to perform the worst. The precision of the semi-automated MeshMonk method was on average 1.97 ± 1.34 mm for the ten landmarks (Figure 4). Compared to this semi-automatic method, the DiffusionNet-based method was found to have significantly better precision for the left exocanthion, endocanthions, nose tip, and cheilions and worse precision for the alares; no significant differences were found for nasion and right exocanthion.

Table 3. The intra-observer variability for all ten landmarks is stated in millimeters ± standard deviation. The intra-observer variability was determined by computing the Euclidean distance between annotations performed in twofold by a single observer.

| Exocanthion | | Endocanthion | | Nasion | Nose tip | Alare | | Cheilion | |
|---|---|---|---|---|---|---|---|---|---|
| *Right* | *Left* | *Right* | *Left* | | | *Right* | *Left* | *Right* | *Left* |
| 0.85 ±0.65 | 0.87 ±0.55 | 0.79 ±0.88 | 0.79 ±0.72 | 1.32 ±1.04 | 0.88 ±0.43 | 1.04 ±0.53 | 0.99 ±0.58 | 0.99 ±0.79 | 0.85 ±0.64 |

Table 4. The interobserver variability is computed by comparing the Euclidean distance between annotations made by three different observers. The Euclidean distances are stated in millimeters ± standard deviation.

| | Exocanthion | | Endocanthion | | Nasion | Nose tip | Alare | | Cheilion | |
|---|---|---|---|---|---|---|---|---|---|---|
| | *Right* | *Left* | *Right* | *Left* | | | *Right* | *Left* | *Right* | *Left* |
| **Observer 1 vs 2** | 1.16 ±0.65 | 1.14 ±0.67 | 1.08 ±0.69 | 0.87 ±0.58 | 1.64 ±0.91 | 1.16 ±0.59 | 1.34 ±0.93 | 1.27 ±0.79 | 0.93 ±0.55 | 0.97 ±0.66 |
| **Observer 1 vs 3** | 1.02 ±0.85 | 0.95 ±0.63 | 1.03 ±0.80 | 1.08 ±0.73 | 1.80 ±1.20 | 1.77 ±0.86 | 1.68 ±1.19 | 1.35 ±0.80 | 1.65 ±1.04 | 1.43 ±1.05 |
| **Observer 2 vs 3** | 1.31 ±0.88 | 1.03 ±0.57 | 0.97 ±0.73 | 1.05 ±0.64 | 2.21 ±1.30 | 2.20 ±1.07 | 1.33 ±0.98 | 1.35 ±0.82 | 1.27 ±0.83 | 1.25 ±0.84 |
| **Average** | 1.16 ±0.80 | 1.04 ±0.63 | 1.03 ±0.74 | 1.00 ±0.66 | 1.88 ±1.17 | 1.71 ±0.96 | 1.45 ±1.05 | 1.32 ±0.80 | 1.28 ±0.88 | 1.22 ±0.88 |

Table 5. The precision of the rough (first DiffusionNet) and refined (second DiffusionNet) is determined by computing the Euclidean distance between the DiffusionNet-predicted and manually annotated landmarks and is stated in millimeters ± standard deviation.

| | Exocanthion | | Endocanthion | | Nasion | Nose tip | Alare | | Cheilion | |
|---|---|---|---|---|---|---|---|---|---|---|
| | *Right* | *Left* | *Right* | *Left* | | | *Right* | *Left* | *Right* | *Left* |
| **Rough predictions** | 2.94 ±2.38 | 2.86 ±1.81 | 2.76 ±2.40 | 2.83 ±2.56 | 1.69 ±1.05 | 1.58 ±0.89 | 2.41 ±1.98 | 2.52 ±1.92 | 3.48 ±3.67 | 3.51 ±2.89 |
| **Refined predictions** | 2.25 ±1.23 | 2.03 ±1.27 | 1.37 ±0.86 | 1.48 ±1.00 | 1.48 ±1.02 | 1.14 ±0.73 | 1.79 ±1.07 | 1.75 ±1.11 | 1.71 ±1.26 | 1.88 ±1.34 |

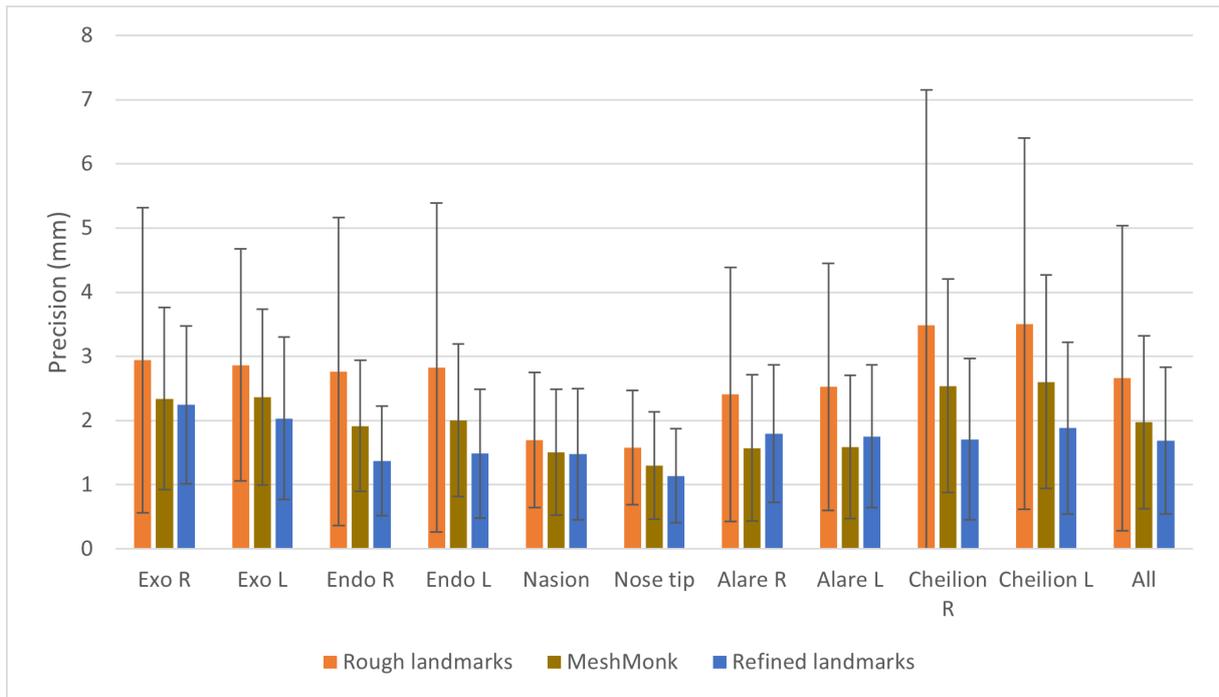

*Figure 4: The precision of the prediction of the rough landmarks (first DiffusionNet), the refined landmarks (second DiffusionNet), and the semi-automated MeshMonk method are visualized for the right exocanthion (Exo R), left exocanthion (Exo L), right endocanthion (Endo R), left endocanthion (Endo L), nasion, nose tip, right alare (Alare R), left alare (Alare L), right cheilion (Cheilion R), and left cheilion (Cheilion L).*

*Table 6: Overview of the accuracy distribution of each landmark as predicted by the complete workflow.*

|  | Percentage of landmarks predicted with a precision within range | | | |
| --- | --- | --- | --- | --- |
|  | < 2 mm | < 3 mm | < 4 mm | < 5 mm |
| **Exocanthion right** | 47% | 77% | 90% | 97% |
| **Exocanthion left** | 56% | 80% | 82% | 97% |
| **Endocanthion right** | 80% | 96% | 99% | 100% |
| **Endocanthion left** | 76% | 92% | 98% | 99% |
| **Nasion** | 77% | 94% | 97% | 99% |
| **Nose tip** | 88% | 98% | 99% | 100% |
| **Alare right** | 62% | 88% | 96% | 99% |
| **Alare left** | 67% | 86% | 96% | 98% |
| **Cheilion right** | 71% | 89% | 95% | 97% |
| **Cheilion left** | 65% | 87% | 94% | 97% |
| **All Landmarks** | 69% | 89% | 96% | 98% |

## Discussion

Soft-tissue cephalometric analysis can be used to objectify the clinical observations on 3D photographs, but manual annotation, the current gold standard, is time-consuming and tedious. Therefore, this study developed a deep learning-based approach for automated landmark extraction from randomly oriented 3D photographs. The performance was assessed for ten cephalometric landmarks: the results showed that the deep-learning-based landmarking method was precise and consistent, with a precision that approximated the inter-observer variability of the manual annotation method. A precision <2 mm, which may be considered a cut-off value for clinical relevance, was seen for 69% of the predicted landmarks (Hsu et al., 2013; Schouman et al., 2015).

In the field of craniofacial surgery, different studies have applied deep-learning models for automated cephalometric landmarking, mainly focusing on 2D and 3D radiographs. Dot et al. used a SpatialConfiguration-Net for the automated annotation of 33 different 3D hard-tissue landmarks from CT images and achieved a precision of $1.0 \pm 1.3$ mm (Dot et al., 2022). An automated landmarking method, based on multi-stage deep reinforcement learning and volume-rendered imaging, was proposed by Kang et al. and yielded a precision of $1.96 \pm 0.78$ mm (Kang et al., 2021). A systematic review by Serafin et al. found a mean precision of 2.44 mm for the prediction of 3D hard-tissue landmarks from CT and CBCT images (Serafin et al., 2023).

Some studies did describe automated algorithms for 3D soft-tissue landmarking on 3D photographs, but these algorithms did not include deep learning models. Baksi et al. described an automated method, involving morphing of a template mesh, for the landmarking of 22 soft-tissue landmarks from 3D photographs that achieved a precision of $3.2 \pm 1.6$ mm (Baksi et al., 2021). An automated principal component analysis-based method, described by Guo et al., achieved an average root mean square error of 1.7 mm for the landmarking of 17 soft-tissue landmarks from 3D photographs (Guo et al., 2013). Even though a direct comparison is infeasible to make due to the difference in landmarks, datasets, and/or imaging modalities, the precision of the proposed workflow is within the same range as these studies.

The effect of landmark choice on the established precision is underlined by the MeshMonk results found in this study. In the original publication by White et al., an average error of 1.26 mm for 19 soft-tissue landmarks. The same methodology was used to establish the precision for the ten landmarks used in this study, and an overall precision of $1.97 \pm 1.34$ mm was found. This finding highlights the difficulty in comparing landmarking precision from literature (White et al., 2019). Compared to the semi-automatic method, the fully-automated workflow yielded significantly improved precision for six landmarks, emphasizing the feasibility of fully-automatically annotating soft tissue landmarks from 3D photos using deep learning.

The proposed workflow uses two successive networks and additional algorithms for alignment and facial segmentation. Advantages of this approach include that the DiffusionNet assures robustness against sampling densities and the HKS settings inherently account for rotational, positional, and scale invariance that may arise between different 3D photography systems. A limitation of the current study is that the workflow was only applied to 3D photographs captured using one 3D photography system. Despite the robust nature of DiffusionNet/HKS, the performance of the workflow might be affected when applied to 3D photographs captured with different hardware. Furthermore, the DiffusionNet models were only trained on spatial features, whereas in the manual annotation process texture information was used. Even though this has the advantage of making the DiffusionNet models insensitive to variations in skin tone or color, landmarks such as the exocanthions, endocanthions, and cheilions could presumably be located more precisely using manual annotation. This would not apply to the landmarks lacking color transitions, such as the nasion and nose tip. Based on these presumptions, the DiffusionNet-based approach might achieve a better precision if texture data of the 3D photographs would be available to the networks.

Another limitation of the proposed workflow arises from the utilization of HKS settings in the initial DiffusionNet, leading to occasional issues with random left-right flipping in the predictions of symmetrical landmarks (e.g., exocanthions). To overcome this challenge, a solution was devised that involved detecting symmetrical landmarks within a single channel. Subsequently, both landmarks were distinguished from each other using a clustering algorithm, followed by a left-right classification based on the midsagittal plane. Although a success rate of 98.6% was achieved using this solution, the workflow failed when the initial DiffusionNet was unable to predict one of the landmarks in the midsagittal plane (nasion, nose tip, or cheilion midpoint). Since this was mainly due to suboptimal quality of the 3D photo, it might be prevented by optimizing image acquisition. For optimal performance of the workflow, it is important to minimize gaps and restrict the depicted area in 3D photos to the face.

Due to its high precision and consistency, the developed automated landmarking method has the potential to be applied in various fields. Possible applications include objective follow-up and analysis of soft-tissue facial deformities, growth evaluation, facial asymmetry assessment, and integration in virtual planning software for 3D backward planning (Memon et al., 2021; Tel et al., 2023). Considering that the proposed DiffusionNet-based approach only uses spatial features, it could be applied on 3D meshes of facial soft tissue that are derived from imaging modalities lacking texture, such as CT, CBCT, or MRI. Nevertheless, further research is necessary to ascertain the applicability of this workflow to these imaging modalities. The fully-automated nature of the workflow also enables cephalometric analysis on large-scale datasets, presenting significant value for research purposes. The position-independency of the workflow might make it suitable for automated landmarking in 4D stereophotogrammetry and give rise to real-time cephalometric movement analysis for diagnostic purposes (Harkel et al., 2020; Shujaat et al., 2014).

**Conclusion**

In conclusion, the effectiveness of a deep learning-based approach for automated landmark extraction from 3D facial photographs was developed and its precision was evaluated. The results showed high precision and consistency in landmark annotation, comparable to manual and semi-automatic annotation methods. Automated landmarking methods offer potential for analyzing large datasets, with applications in orthodontics, genetics, and craniofacial surgery and in emerging new imaging techniques like 4D stereophotogrammetry.

**Author contributions**
**Bo Berends:** Conceptualization, Methodology, Software, Validation, Writing – Original Draft **Freek Bielevelt:** Conceptualization, Methodology, Software, Validation, Writing – Original Draft **Ruud Schreurs:** Conceptualization, Writing – Review & Editing **Shankeeth Vinayahalingam:** Writing – Review & Editing **Thomas Maal:** Supervision, Lab Management **Guido de Jong:** Conceptualization, Software, Formal Analysis, Supervision.

**Declaration of Competing Interest**
There are no conflicts of interest to declare.
**Ethical Approval and waiver list:** CMO 2007/163; ARB NL 17934.091.07; RUMC CMO 2019-5793

**Funding:** This research did not receive any specific grant from funding agencies in the public, commercial, or not-for-profit sectors.

**Data availability:** Coded scripts are available within the following GitHub repository:
https://github.com/rumc3dlab/3dlandmarkdetection/

# Appendix

*Appendix Table 1: The parameters set used for the configuration of the MeshMonk algorithm.*

| Parameters | Values |
|---|---|
| **Rigid Registration** | |
| *Number of iterations* | 30 |
| *Correspondence neighbor number* | 3 |
| *Correspondence flag threshold* | 0.90 |
| *Correspondence symmetric* | Yes |
| *Correspondence equalize* | No |
| *Use scaling* | Yes |
| *Inlier kappa* | 4.00 |
| *Inlier use orientation* | Yes |
| *Floating boundary* | Yes |
| *Target boundary* | Yes |
| *Target badly shaped triangles* | Yes |
| *Triangle size Z factor* | 6.00 |
| *Target up sample* | No |
| **Non-rigid registration** | |
| *Number of iterations* | 80 |
| *Correspondence neighbor number* | 3 |
| *Correspondence flag threshold* | 0.90 |
| *Correspondence symmetric* | Yes |
| *Correspondence equalize* | No |
| *Inlier kappa* | 12.00 |
| *Inlier use orientation* | Yes |
| *Floating boundary* | Yes |
| *Target boundary* | Yes |
| *Target badly shaped triangles* | Yes |
| *Triangle size Z factor* | 6.00 |
| *Target upsample* | Yes |
| *Inlier use weights* | No |
| *Transform sigma* | 3.00 |
| *Viscous iteration start* | 200 |
| *Viscous iteration end* | 1 |
| *Elastic iteration start* | 200 |
| *Elastic iteration end* | 1 |
| *Transform neighbors* | 80 |